\documentclass[lettersize,journal,twoside,article]{IEEEtran}
\usepackage{amsmath,amsfonts}
\usepackage{algorithmic}
\usepackage{algorithm}
\usepackage{array}
\usepackage[caption=false,font=normalsize,labelfont=sf,textfont=sf]{subfig}
\usepackage{textcomp}
\usepackage{stfloats}
\usepackage{url}
\usepackage{verbatim}
\usepackage{graphicx}
\usepackage{cite}
\usepackage{multirow}
\usepackage{hyperref}
\usepackage[all]{hypcap}
\usepackage{mathbbol}

\hypersetup{
    colorlinks=true,
    linkcolor=blue,
    citecolor=blue,
    urlcolor=blue,
}

\usepackage{tikz}
\usetikzlibrary{shapes.geometric, arrows, fit}

\tikzstyle{input} = [rectangle, rounded corners, 
minimum width={width("Award text description input")+2pt}, 
minimum height=1cm,
text centered, 
draw=black, 
fill=red!30]

\tikzstyle{output} = [rectangle, rounded corners, 
minimum width={width("Technology")+2pt}, 
minimum height=0.5cm,
text centered, 
draw=black, 
fill=green!50]

\tikzstyle{processedInput} = [rectangle, rounded corners, 
minimum width=3cm, 
minimum height=1cm,
text centered, 
draw=black, 
fill=red!50]

\tikzstyle{encoder} = [trapezium, rounded corners, 
minimum width=1cm, 
minimum height={width("Encodern")+10pt}, text centered, 
draw=black, fill=blue!30]

\tikzstyle{concat} = [rectangle, rounded corners, 
minimum width=7cm, 
minimum height=1cm,
text centered, 
draw=black, 
fill=yellow!50]

\tikzstyle{feature} = [rectangle, rounded corners, 
minimum width=10.5cm, 
minimum height=1cm,
text centered, 
draw=black, 
fill=yellow!50]

\tikzstyle{cluster} = [diamond, rounded corners, 
minimum width=3cm, 
minimum height=1cm, 
text centered, 
draw=black, 
fill=orange!50]

\tikzstyle{fc1} = [trapezium, rounded corners, 
minimum width=3cm, 
minimum height=1cm, text centered, 
draw=black, fill=blue!45]

\tikzstyle{fc2} = [trapezium, rounded corners, 
minimum width=3cm, 
minimum height=3cm, text centered, 
draw=black, fill=blue!60]

\tikzstyle{arrow} = [thick,->,>=stealth]

\begin{document}

\title{Predicting Multi-Type Talented Students in Secondary\\ School Using Semi-Supervised Machine Learning}

\author{Xinzhe Zheng, Zhen-Qun Yang, Jiannong Cao*, Jiabei Cheng
\thanks{Manuscript created August 2024; This work is supported by Hong Kong Jockey Club Charities Trust (Project S/N Ref.: 2021-0369), and the Research Institute for Artificial Intelligence of Things, The Hong Kong Polytechnic University. The opinions expressed here are entirely that of the author. No warranty is expressed or implied. User assumes all risk.

Xinzhe Zheng, Zhen-Qun Yang, Jiannong Cao, Jiabei Cheng
are with the Department of Computing, The Hong Kong Polytechnic University, Hong Kong, China
(e-mail:
xzhengbj@connect.ust.hk; 
zq-cs.yang@polyu.edu.hk; 
jiannong.cao@polyu.edu.hk; 
bettybei310@gmail.com).
}}

\maketitle
\begin{abstract}
Talent identification plays a critical role in promoting student development. However, traditional approaches often rely on manual processes or focus narrowly on academic achievement, and typically delaying intervention until the higher education stage. This oversight overlooks diverse non-academic talents and misses opportunities for early intervention. To address this gap, this study introduces TalentPredictor, a novel semi-supervised multi-modal neural network that combines Transformer, LSTM, and ANN architectures. This model is designed to predict seven different talent types—academic, sport, art, leadership, service, technology, and others—in secondary school students within an offline educational setting.
Drawing on existing offline educational data from 1,041 local secondary students, TalentPredictor overcomes the limitations of traditional talent identification methods. By clustering various award records into talent categories and extracting features from students' diverse learning behaviors, it achieves high prediction accuracy (0.908 classification accuracy, 0.908 ROCAUC). This demonstrates the potential of machine learning to identify diverse talents early in student development. 

\end{abstract}


\section{Introduction}
\IEEEPARstart{T}{alent} is a critical component in human society. It is indispensable to the development of societies and the competitiveness of countries. Last but not least, talent is always in high demand. Thus, nurturing talent is the top priority for every part of the earth, and in it, talent identification is the foundation, as you must have a target individual to nurture talent. Traditional talent identification aims to give students tests that exceed their current level. For example, give grade eight students college admissions tests and use the result of the tough test as a talent score. Other methods are marking those students who took courses above their current level or received outstanding awards as talented \cite{talentSearch}.

However, traditional talent identification is manual and time-consuming for teachers in the era of artificial intelligence (AI). As AI has advanced in recent years, its application has become ubiquitous in various industries, including but not limited to finance, healthcare, and education, inspiring innovation and revolutionizing them. In particular, the application of AI in education has developed into a new research field: big data in education \cite{wei2012mining}, which become an emerging research recently \cite{Nayak_2022, Mishra_2024, Daniel_2021}. In big data in education, AI has been applied to education for academic performance prediction of students, which automates the tedious process and saves teachers much time in assessing student performance. However, big data in academic performance prediction of students only covers academic talent prediction. Research related to other talent types is not found with the best effort in surveys \cite{Nayak_2022, Mishra_2024, Daniel_2021}, indicating big data research on multi-type talent identification has not been explored yet. In terms of targeted level of education, most existing research about academic performance prediction using AI is focused on higher education only. Related research on secondary school education can barely be found with the best effort. Lastly, note that the student population has a more talented proportion higher up in the education hierarchy and a less talented proportion lower down in the education hierarchy. So talent identification processes are more likely to discover previously missed potential talents at lower levels of the education hierarchy. Hence, research in identifying students’ talents at an earlier stage with machine learning is more beneficial than at a later stage. 

\IEEEpubidadjcol
To address the issue, this research focuses on the multi-type talent identification problem in secondary school, exploring how to solve the problem using machine learning. We want to answer the following three research questions:

1) In the offline educational environment, to what degree can machine learning identify a variety of student talents? 

2) Is it possible to identify students' diverse talents at an earlier stage with machine learning?

3) What factors are associated with students' various talents?

The main contributions of this research are as follows:

1) A novel semi-supervised multimodal model (TalentPredictor) was constructed for multi-type talent identification in secondary school offline education settings. The model is trained and evaluated on a real-world dataset with 1041 students.

2) TalentPredictor demonstrates accurate talent forecasting by predicting future manifestations of student talents one semester in advance using existing data, surpassing conventional test-dependent approaches. 

3) Analysis of the key relevant factors underlying the seven types of student talent was conducted.

This article is structured as follows. Section II presents previous research on the application of big data in academic performance prediction and popular machine learning models. Section III describes the details of TalentPredictor. Section IV presents the model performance results and comparisons between algorithms and metrics. Section V discusses the results in depth, analyzing the model’s interpretability, and potential ethical implications, and addresses the research questions. Finally, Section VI summarizes the result, draws conclusions, and suggests possible future research.

\section{Related Work}
In this section, we first provide an overview of existing works on the application of big data to academic performance prediction. This helps capture the current research landscape and identify the state-of-the-art approaches. Next, we introduce several popular machine learning algorithms commonly used in big data applications to build a general understanding of the current big data technologies and their capabilities.
Finally, we explore the limitations of technology-driven approaches in talent identification, emphasizing the gaps in effectively addressing the identification of multi-type talents at an early stage.

\subsection{The application of big data in academic performance prediction.}
The application of big data in academic performance prediction has gained a lot of research interest in recent years. It focuses on analyzing a large amount of data in educational databases regarding academic performance. To gain insight into the data, statistics and machine learning methods were used to assist the teacher in performing better at their job. To explore multiple typed talented students' predictions, existing literature on student academic talented prediction, such as academic performance prediction using machine learning, was reviewed in detail.

The majority of existing work is predicting the academic performance of university students. Grade point average (GPA) scores, levels at low, middle, and high, of undergraduates have been used as common prediction targets. M. Mohammadi et al. \cite{8669085} utilize K-Nearest Neighbors (KNN), Naïve Bayes Classifier (NB), and Decision Tree (DT) algorithms in their study. Their best result is an accuracy of 0.546 using KNN on the GPA level prediction. GPA level prediction was also explored by applying features selection first, followed by a classifier for prediction \cite{punlumjeak2017big}. The research uses Chi-square,  Pearson  Correlation,  and  Mutual information as feature selection methods. Next, they use DT and Artificial Neural Networks (ANN) as classifiers, and the best accuracy achieved is 0.906. Another research also explored a similar question. In it, the GPA of university students has been predicted using logistic regression (LR), DT, and support vector machine (SVM). And the best accuracy reached is 0.960 \cite{Lukman_2024}. The academic pathways of high school students have also been predicted using DT, Xtreme Gradient Boosting (XGB), Random Forest (RF), NB, NB with bagging, and ANN \cite{Abdalkareem_2024}. The best result has an accuracy of 0.990, precision, recall, and F1-score, and it achieved the highest AUC metric by scoring 1.000. The academic performance of graduating college students has been predicted by \cite{Estandarte_10544915} using DT, LR, RF, SVM, and NB with the best accuracy of 0.910. The academic performance in secondary school has been predicted by \cite{Hoyos_data9070086}. They use ANN and SVM with the best result of an average accuracy of 0.880 between different subjects of prediction. The final exam grades of undergraduate students have been predicted by \cite{yac_2022_educational}. They used an assembled model of RF, SVM, LR, KNN, NB, and ANN resulting in the best classification accuracy of 0.750.

To conclude, previous research on student academic performance or status prediction uses only one model or assemble of models for predictions. Models used are all either statistical methods or sallow networks that can be interpreted by humans easily. No existing approach uses a semi-supervised multimodal model with deep learning (DL) to tackle the problem. However, DL has a significant advantage over those models, the automatic and scalable feature extraction. This allows DL to perform well without the tremendous amounts of feature engineering required by statistical methods and sallow networks. This also enables DL to be deployed to new schools with new data without a significant amount of feature engineering overhead. Lastly, related research on the prediction of multi-type talented students in secondary school education was not found under the best effort. 

\subsection{Machine learning methods to predict talented students}
Machine learning is a pivotal subfield of AI that emphasizes the development of algorithms and statistical models to recognize patterns in data. Famous natural language processing (NLP) tools such as GPT and Llama are all based on ML, proving the power of ML. In this section, popular ML algorithms useful for predicting talented students will be introduced, including clustering algorithms, the transformer, Bidirectional Encoder Representations from Transformers (BERT), and Long Short-Term Memory (LSTM). Their intuition behind it is introduced briefly in this section, formal mathematical definitions can be found in the next section.

Clustering is a type of unsupervised learning, and the biggest advantage of unsupervised learning compared to supervised learning is the absence of human labeling. So it is suitable for tasks that are difficult or impossible to be human-labeled. Next, the twelve most popular clustering algorithms will be introduced, namely: Affinity Propagation \cite{affinity_propagation}, Agglomerative Clustering using Average Linkage, Agglomerative Clustering using Ward Linkage, Birch \cite{birch}, DBSCAN \cite{DBSCAN}, HDBSCAN \cite{HDBSCAN}, Gaussian Mixture \cite{gaussian_mixture}, K-Means, Mean Shift \cite{mean_shift}, Mini Batch k Means, OPTICS \cite{optics}, and Spectral Clustering \cite{spectral_clustering}. 

Affinity Propagation is based on passing messages between data points as the name of the original paper 'Clustering by Passing Messages Between Data Points' suggests \cite{affinity_propagation}. It uses this message to locate cluster centers and assign data to the cluster centers of the output cluster. Agglomerative Clustering is a bottom-up hierarchical clustering, that uses the linkage method to decide which two sets to merge. The average linkage simply uses averages of the distance of data in each of the two sets and merges the smallest out of all. While ward linkage uses variance and merges the two sets have the minimum variance. Birch is specialized for large datasets and converges faster compared with other algorithms. It converts data to a balanced tree named CF tree, then condenses the tree, and finally applies hierarchical clustering to get the cluster. DBSCAN is a clustering using density between each data point. It clusters high-density data points to the same group. Roughly speaking, the cluster is formed by firstly connecting all pairs of points within a certain distance, eps, given by the user, and points connected are in the same group. Finally, eliminating those groups with too few points and the result is the cluster. OPTICS can be considered a generalization of DBSCAN that relaxes the eps requirement from a single value to a value range \cite{scikit-learn}. So it can produce clusters of different densities at one go. HSBDCAN is a variation based on DBSCAN by using hierarchical clustering and OPTICS. It is a DBSCAN with an algorithm finding an optimal distance eps for the cluster. K-Means and Mini Batch K-Means is the classic clustering using centroid. The algorithm assigns data points to centroids then updates centroids to the mean of assigned data points and repeats this process until centroids remain unchanged. Gaussian Mixture fit probability distributions over data points using expectation-maximization (EM) algorithm to form clusters. Mean Shift clustering aims to discover blobs in a smooth density of samples. It is a centroid-based algorithm, which works by updating candidates for centroids to be the mean of the points within a given region \cite{scikit-learn}. Spectral Clustering performs a low-dimension embedding of the affinity matrix between samples, followed by clustering, e.g., by K-Means, of the components of the eigenvectors in the low dimensional space \cite{scikit-learn}.

LSTM \cite{lstm_1997} is a supervised recurrent neural network (RNN) model for time series. LSTM solves the gradient explosion or vanishing problem in RNN by using an efficient, gradient-based algorithm for an architecture enforcing constant error flow through internal states of special units. Roughly speaking, this neural network can recognize the general trend of all data fed in long-term memory while storing the short-term memory comes at the last in input without altering the general trend.

Transformer \cite{transformer_2017} is a supervised NLP deep learning neural network model for machine translation. It employs only attention layers in the neural network. To translate, it first produces an embedding of the source language using an encoder and decodes the embedding into the target language using a decoder, where the embedding is a high-dimensional numerical vector capturing useful features from the input. Noticing that the intermediate embedding can represent the semantics meaning of natural language in a numerical form precisely, BERT \cite{bert} started to use the encoder of the transformer to train embedding to represent natural language unsupervised. It uses fill-in-the-blank and next-sentence prediction as training tasks to replace the original translation task. Both of the tasks don't need human labels, instead, comparing masked text and shuffled paragraphs with the original can serve as the ground truth labels, thus the training becomes unsupervised. When trained on a very large amount of text data, BERT can produce high-quality text embedding for general content. One could further fine-tune the pre-trained model on a specific downstream task so the model can produce more precise embedding on the given downstream task. 

\subsection{Limitations of the current technology-driven approaches in the context of talent identification}

Despite significant advancements in big data and machine learning, current talent identification methods exhibit two significant shortcomings:

\textbf{1. Narrow Talent Recognition.}
Most methods prioritize academic achievement as the primary indicator of potential. Traditional metrics including GPA and grades, dominate talent assessment \cite{ 8669085, punlumjeak2017big, Lukman_2024, Estandarte_10544915, Hoyos_data9070086, yac_2022_educational}. This narrow focus often overlooks diverse talents that do not align with traditional academic measures.
For instance, Ammar predicted undergraduate performance based on past academic records  \cite{Almasri}. Similarly, Noel clustered university students based on their grades \cite{10.1007/978-981-32-9563-6_19}, while Alaa employed decision trees and questionnaires to forecast course failures among Computer Science and Information Technology students \cite{Hamoud2018}. Other studies, such as those by Abdelaziz and Raza, also prioritized academic indicators to classify student talents and predict performance \cite{abdelhamid2022machine, 8510600}.
Furthermore, some research, such as that conducted by Ishwank and Ramla, has focused solely on academic metrics, even utilizing electroencephalogram data to predict test performance \cite{Ghali2019UsingEF}. In instances where real data collection posed challenges, Hodges resorted to using artificially generated data to train simple artificial neural networks and logistic regression models for talent prediction  \cite{hodges2019machine}.

Even in studies aimed at identifying talents in non-academic fields, such as sports, researchers often substitute standard academic exams with specially designed tests while employing similar statistical methods for prediction \cite{sports10060081}. This trend inadvertently reduces the complex, multidimensional concept of talent to mere exam scores. Such over-reliance on academic results creates systemic blind spots, leaving untapped potential in students unrecognized if they do not engage in relevant courses or assessments. Therefore, there is an urgent need to develop alternative methods that are less dependent on traditional exam outcomes. Drawing from the research approach presented in the paper \cite{wei2008fusing}, we can integrate different evaluation metrics to achieve a more comprehensive and nuanced recognition of talent.

\textbf{2. Delayed Identification.} 
The second flaw is that talent identification predominantly begins at the university level, neglecting earlier stages, such as high school. This is a time when students are forming their identities and developing essential character traits.  This delay limits the ability to discover and nurture various talents before students enter higher education.

To address these gaps, this research introduces a deep learning model called TalentPredictor, designed to identify various types of talents from multiple perspectives at an earlier stage in the educational journey.

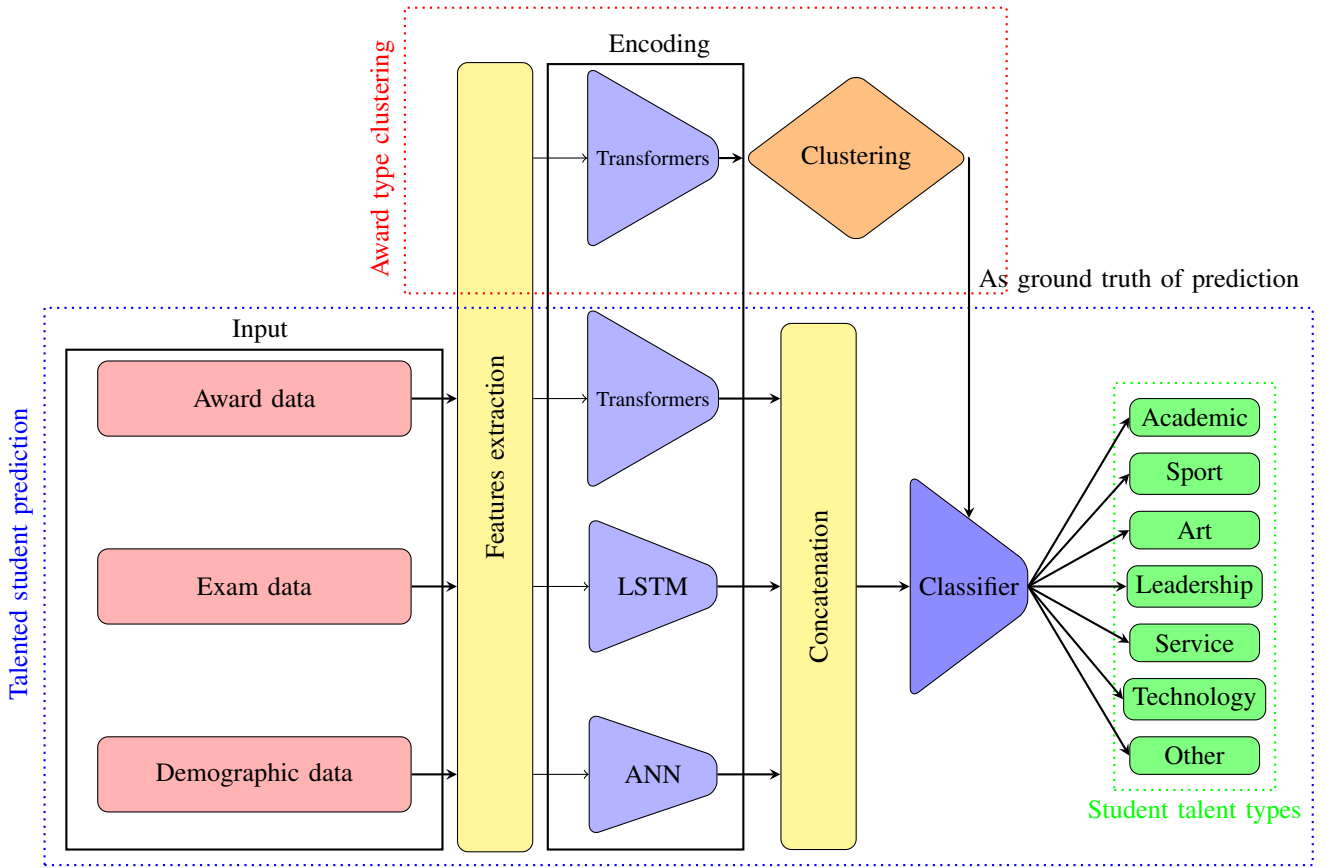
\begin{figure*}
\centering
\begin{tikzpicture}[node distance=2cm]
\tikzset{
    trapezium stretches=true
}
\node (awardText) [draw=none] {};
\node (onehotIn) [input, below of=awardText, yshift=-1.2cm] {Award data};
\node (encode_oh) [encoder, shape border rotate=270, right of=onehotIn, xshift=3.3cm] {\footnotesize{Transformers}};
\node (seqIn) [input, below of=onehotIn, yshift=-0.5cm] {Exam data};
\node (encode_seq) [encoder, shape border rotate=270, right of=seqIn, xshift=3.3cm] {LSTM};
\node (txtIn) [input, below of=seqIn, yshift=-0.5cm] {Demographic data};
\node (transformer1) [encoder, shape border rotate=270, right of=awardText, xshift=3.3cm] {\footnotesize{Transformers}};
\node (cluster) [cluster, right of=transformer1, xshift=0.7cm] {Clustering};
\node (transformer2) [encoder, shape border rotate=270, right of=txtIn, xshift=3.3cm] {ANN};
\node (concat) [concat, rotate=90, right of=encode_seq, xshift=-2cm, yshift=-2.2cm] {Concatenation};

\node (fc1) [fc1, shape border rotate=270, right of=concat] {Classifier};

\node (output1) [output, right of=fc1, xshift=1cm, yshift=2.25cm] {Academic};
\node (output2) [output, right of=fc1, xshift=1cm, yshift=1.5cm] {Sport};
\node (output3) [output, right of=fc1, xshift=1cm, yshift=0.75cm] {Art};
\node (output4) [output, right of=fc1, xshift=1cm] {Leadership};
\node (output5) [output, right of=fc1, xshift=1cm, yshift=-0.75cm] {Service};
\node (output6) [output, right of=fc1, xshift=1cm, yshift=-1.5cm] {Technology};
\node (output7) [output, right of=fc1, xshift=1cm, yshift=-2.25cm] {Other};

\node (fea_extraction) [feature, rotate=90, right of=onehotIn, xshift=-2.78cm, yshift=-3.2cm] {Features extraction};

\draw[draw,red,thick,dotted] (2,-1.8) rectangle (10,2) node [left, rotate=90, xshift=-0.3cm, yshift=8.3cm, red] (BoxNode1) {Award type clustering};
\node[draw,blue, thick,dotted,inner sep=0.7cm,label={[rotate=90, xshift=2.5cm, yshift=0.3cm, blue]left:Talented student prediction},fit=(onehotIn) (output1) (output7) (txtIn)] {};
\node[draw,green, thick,dotted,inner sep=0.2cm,label={[green]below:Student talent types},fit=(output1) (output7)] {};
\draw[thick, black] (-2.5,-9.2) rectangle (2.5,-2.55) node [above, xshift=-2.4cm, pos=0.995] (BoxNode1) {Input};
\draw[thick, black] (3.9,-9.2) rectangle (6.5,1.25) node [above, xshift=-1.1cm, pos=0.995] (BoxNode2) {Encoding};

\draw [arrow] (transformer1) -- (cluster);
\draw [arrow] (onehotIn.east) -- (onehotIn-|fea_extraction.north);
\draw [arrow] (encode_oh.east) -- (encode_oh-|concat.north);
\draw [arrow] (seqIn.east) -- (seqIn-|fea_extraction.north);
\draw [arrow] (encode_seq.east) -- (encode_seq-|concat.north);
\draw [arrow] (txtIn.east) -- (txtIn-|fea_extraction.north);
\draw [arrow] (transformer2.east) -- (transformer2-|concat.north);
\draw [<-] (transformer1) -- (transformer1-|fea_extraction.south);
\draw [<-] (encode_oh) -- (encode_oh-|fea_extraction.south);
\draw [<-] (encode_seq) -- (encode_seq-|fea_extraction.south);
\draw [<-] (transformer2) -- (transformer2-|fea_extraction.south);

\draw [arrow] (concat) -- (fc1);
\draw [arrow] (fc1.east) -- (output1.west);
\draw [arrow] (fc1.east) -- (output2.west);
\draw [arrow] (fc1.east) -- (output3.west);
\draw [arrow] (fc1.east) -- (output4.west);
\draw [arrow] (fc1.east) -- (output5.west);
\draw [arrow] (fc1.east) -- (output6.west);
\draw [arrow] (fc1.east) -- (output7.west);
\draw [arrow] (cluster.east) -| (fc1.north) node [pos=.67, right] (TextNode) {As ground truth of prediction};
\end{tikzpicture}
\caption{The architecture of our TalentPredictor model. Raw input data is first processed by the feature extraction. Next, corresponding features are fed into corresponding encoders. Awards type clustering group award description embedding to give the award a talent type and prepare it as the ground truth for the talented student prediction training. Talented student prediction concatenates outputs from the relevant encoder and uses a final classifier to predict the talented fields of a student.}
\label{architecture}
\end{figure*}

\section{Methods}
This section provides detailed definitions of seven types of talents in secondary school students, describes the dataset used, and presents the novel semi-supervised multi-modal model, TalentPredictor, proposed in this research.

\subsection{The Definitions of Seven Types of Talents in Secondary School Students}
Talented, not to be confused with gifted refers to those individuals who come with superior inborn ability, such as the ability to learn a certain subject faster, among the top few percent of the whole population. While talented refers to those people with honed skills in certain subjects \cite{_2015_supporting}. This research is targeted to identify talented students instead of gifted students in secondary school education. To quantify talented students, local secondary school teachers were interviewed, and the discussion led to the conclusion of the following definition: A student with at least merit awards from activities involving participant size at least of the whole school in a subject indicates the student is talented in that subject. Because an awardee must have honed skill for a certain subject to be awarded accordingly. Lastly, the talented types are divided into the following seven: Academic, Sport, Art, Leadership, Service, Technology, and Other. The seven types of talent are concluded by local secondary school teachers based on their experience and following Marland Definition \cite{talent_definition}: 

Gifted and talented children capable of high performance include those with demonstrated achievement and/or potential ability in any of the following areas, singly or in combination: (1) General intellectual ability, (2) Specific academic aptitude, (3) Creative or productive thinking, (4) Leadership ability, (5) Visual and performing arts, (6) Psychomotor ability (i.e., superior athleticism). 

\subsection{Dataset}
To address the two shortcomings of existing talent identification methods discussed earlier, our approach does not rely on specially designed exams for each talent category (for instance, art talent identification would require exams specifically tailored for art). Instead, we utilize the existing offline educational dataset that easy to capture from the schools. Our experiment focuses on a secondary school with 1,041 students.

By moving beyond the narrow focus on grades, we acknowledge that talents extend beyond traditional metrics. Our dataset includes quantitative scores only in three core subjects: English, Chinese, and Mathematics, along with demographic data and qualitative insights into students’ learning behaviors, such as classroom performance, attendance, participation in extracurricular activities (ECA), and involvement in diverse competitions.

This multifaceted approach enables us to identify talents from multiple perspectives, effectively addressing the issue of delayed identification by initiating the evaluation process at an earlier stage in students' educational journeys. Ultimately, our goal is to uncover diverse talents that are often overlooked by conventional methods as early as possible, thereby facilitating a more holistic understanding of student potential.

\subsection{TalentPredictor}
The TalentPredictor mainly consists of three parts: 1) \textit{feature extraction}, 2) \textit{encoding}, and 3) \textit{prediction}.

\subsubsection{Feature Extraction}
After data is collected and labeled, useful features will be extracted and processed to the correct data format before feeding into the model. Firstly, award records will be clustered. This uses the text description of the awards to group awards into seven different groups and uses ground truth for evaluation. The purpose of the cluster is to act as an auto labeler, so TalentPredictor can also apply to data without human labels on award records. For the prediction part, exams, awards, and demographic data of students are used as input and separated into four types of data: \textbf{text data, sequential data, discrete data, and numerical data}. Text data and numerical data are self-explainable. Sequential data is time series data, for example, the exam score of a student in order of occurrence. Discrete data are discontinued types, such as the sex of the student. As for the prediction target, the talent type of awards will be used. If a student has award records of certain talent types, the student is labeled as talented in those fields. Lastly, data augmentation is employed on sequential data. The data augmentation is as such: All sequential data of a student have length \(t\) i.e. all data is in the form of \(\{a_1, a_2, \dots, a_t\}\). Randomly cut the last \(k\) elements, \(\{a_{t-k+1}, a_{t-k+2}, \dots, a_{t}\}\), where \(k\) ranged from 0 to \(t-1\). i.e. only use the first \(t-k\), \(\{a_1, a_2, \dots, a_{t-k}\}\), data to predict whether student is talented at time \(t\), when \(t\) is the latest time student data have been updated. Only sequential data are augmented, since other data are constant throughout the whole high school life of students and will induce meaningless noise. Such augmentation is applied on each sequential data with random \(k\) for each epoch, similar to traditional data augmentation approaches in image classification \cite{Ansel_PyTorch_2_Faster_2024}. Such an approach increases the amount of training data and improves the generalizability of the model without slowing down the training.

\subsubsection{Encoding}
The semi-supervised multimodal model (fig.\ref{architecture}) is constructed with a clustering algorithm, transformer, and LSTM. It is called semi-supervised, as it combines unsupervised and supervised learning.

The encoding has two parts. The first part is the unsupervised part, which uses clustering to group awards into seven different talent types (Academic, Sport, Art, Leadership, Service, Technology, Other). Clustering explored 12 different clustering algorithms in the related work implant in \cite{scikit-learn} and only the best are used in the final implementation. In this part, the transformer pre-trained using a variation of BERT designed for both Chinese and English \cite{cui-etal-2021-pretrain} was used to get the embedding of the description of awards. Finally, clusterings were applied to the embedding of all awards and partition awards into seven groups as in Fig \ref{architecture}. This part is crucial as it gives ground truth for the training and quantifies individual awards as trainable data. 

The second part is the supervised prediction part. It uses the result from the first part to add the prediction target of the training, students' talented type. If students have talented awards records in certain talented types then the student is talented in those types. After preprocessing data, LayerNorm \cite{ba2016layernormalization} will normalize data first, then different types of data are fed into the corresponding encoder. The encoders for sequential data, text data, and discrete data are LSTM, transformer, and ANN respectively, and encoders of different sizes have been experimented with. Only two of the best encoders are reported, namely Raw Encoder and One Encoder. The Raw Encoder uses LSTM with a hidden size of 20 and 2 recurrent layers, hfl/chinese-bert-wwm in \cite{cui-etal-2021-pretrain} for the transformer and identical layer for ANN. The outputs of all encoders are concatenated into one, then normalized by LayerNorm, and finally consumed by the final classifier for talented student prediction. One Encoder is modified based on the Raw Encoder such that all encoders will output embedding of size 1. Finally, all size 1 embedding are concatenated into one, then normalized by LayerNorm, and input to the final classifier as in Fig \ref{architecture}.

Since LSTM and transformer are not popular in big data education, their mathematics definitions are stated as follows to make the paper self-contained. According to LSTM implementation of PyTorch: LSTM is an RNN where the base recurrent unit is an LSTM cell, and the cell is defined by EQ. \ref{lstm_1}-\ref{lstm_6}. In it, \(h_t\) is the hidden state at time \(t\), \(c_t\) is the cell state at time \(t\), \(x_t\) is the input at time \(t\), \(h_{t-1}\) is the hidden state of the layer at time \(t-1\) or the initial hidden state at time \(0\), and \(i_t, f_t, g_t, o_t\) are the input, forget, cell, and output gates, respectively. \(W\) and \(b\) are learnable parameters. \(\sigma\) is the sigmoid function, tanh is the hyperbolic tangent function, and \(\odot\) is the Hadamard product \cite{Ansel_PyTorch_2_Faster_2024}.

In a multilayer LSTM, the input \(x_{t}^{(l)}\) of the \(l^{\text{th}}\) layer \((l \ge 2)\) is the hidden state \(h_t^{(l-1)}\) of the previous layer multiplied by the dropout \(\delta_t^{(l-1)}\) where each \(\delta_t^{(l-1)}\) is a Bernoulli random variable which is 0 with probability defined by the user \cite{Ansel_PyTorch_2_Faster_2024}.

\begin{align} 
\label{lstm_1}
i_t = \sigma(W_{ii}x_t + b_{ii} + W_{hi}h_{t-1} + b_{hi}) \\
\label{lstm_2}
f_t = \sigma(W_{if}x_t + b_{if} + W_{hf}h_{t-1} + b_{hf})\\
\label{lstm_3}
g_t = \text{tanh}(W_{ig}x_t + b_{ig} + W_{hg}h_{t-1} + b_{hg})\\
\label{lstm_4}
o_t = \sigma(W_{io}x_t +b_{io} + W_{ho}h_{t-1} + b_{ho})\\
\label{lstm_5}
c_t = f_t \odot c_{t-1} + i_t \odot g_t\\
\label{lstm_6}
h_t = o_t \odot \text{tanh}(c_t)
\end{align}

To construct an LSTM network, multiple layers of LSTM cells connect in sequence, and time series data is inputted in order of timestamp \(t\).

Transformer is a supervised NLP deep learning neural network utilizing multi-head attention mainly (EQ. \ref{transform_att_1}-\ref{transform_att_3}), where \(W\) is the learnable parameters. Transformer predicts the next word in the sentence of the target language one after another given the source language using encoder-decoder architecture. The encoder takes input and the decoder produces output, the difference between them is decoder has an extract-masked multi-head attention layer. The masked multi-head attention is setting parameters to \(-\inf\) for those not yet seen in encoder input \cite{transformer_2017}.
\begin{align}
\label{transform_att_1}
\text{Attention}(Q, K, V) = \text{softmax}(\frac{QK^T}{\sqrt{d_k}})V\\
\label{transform_att_2}
\text{MultiHead}(Q, K, V) = \text{Concat}(\text{head}_1,\dots,\text{head}_h)W^o\\
\label{transform_att_3}
\text{where head}_i = \text{Attention}(QW_i^Q, KW_i^K, VW_i^V)
\end{align}

The other layers of the model include add and norm. LayerNorm is the layer normalization in \cite{ba2016layernormalization}, while Sublayer(\(x\)) is the function implemented by the layer just before add and norm. i.e. the layer uses \(x\) as the input \cite{transformer_2017}.

\begin{equation}
\label{transform_addnorm}
\text{Add\&Norm}(x) = \text{LayerNorm}(x+\text{Sublayer}(x))
\end{equation}

Feed forward networks \cite{transformer_2017}:

\begin{equation}
\label{transform_ffn}
\text{FFN}(x) = \text{max}(0, xW_1 + b_1)W_2 + b_2
\end{equation}

Positional encoding as in EQ. \ref{transform_pe_1}-\ref{transform_pe_2} where \(pos\) is the text position and \(i\) is the dimension \cite{transformer_2017}:

\begin{align}
\label{transform_pe_1}
PE_{(pos,2i)} = sin(pos/10000^{2i/d_{\text{model}}})\\
\label{transform_pe_2}
PE_{(pos,2i+1)} = cos(pos/10000^{2i/d_{\text{model}}})
\end{align}

And embeddings and softmax \cite{transformer_2017}. Softmax is as EQ. \ref{transform_softmax}, while embeddings is a linear layer, i.e. one layer ANN. The embedding will convert the input tokens and output tokens to the size of the model input. The tokens refer to plain text converted into vectors through a tokenizer such as Bag-of-Words \cite{mikolov2013efficientestimationwordrepresentations}.

\begin{equation}
\label{transform_softmax}
\text{softmax}(\vec{x})_i = \frac{e^{x_i}}{\sum_{j=1}^{K}e^{x_j}}
\end{equation}

Finally, the transformer is finished by connecting all the layers as follows. For the encoder, layers are in sequence of positional embedding, multi-head attention, add \& norm, FFN, add \&norm. For the decoder, layers are in sequence of positional embedding, masked multi-head attention, add \& norm, multi-head attention, add \& norm, FFN, add \&norm. The encoder is connected to the decoder from the final layer of the encoder to the multi-head attention of the decoder. The input layers for both the encoder and decoder are the same, both use a tokenizer followed by embedding. The input for the encoder will be source language, while the input for the decoder will be described later. The output of the transformer is the decoder connected to embedding followed by softmax. The final output is the token with the highest probability in the softmax output. The input for the decoder will be the predicted token from all previous iterations concatenated sequentially. For the first iteration, it will use a special input to indicate it is the first. Lastly, the decoder and encoder can stack N time, i.e. total N encoder connected end to end, and the same for the decoder. This stacking can boost the performance of the transformer model.

After understanding the architecture of TalentPredicton and the mathematics principle behind the transformer \cite{transformer_2017} and LSTM \cite{lstm_1997}, TalentPredicton can be trained. Lastly, models are implemented using PyTorch \cite{Ansel_PyTorch_2_Faster_2024}.

\subsubsection{Prediction}
The final classifier is a fully connected layer with a sigmoid layer and outputs seven numbers \(x\), where \(\{x \in \mathbb{R} | 0 < x < 1\}\). \(x\) represents the confidence that the student is talented in each of the seven talent types, 0 refers to not talented at all while 1 refers to very talented.

\section{Results and Discussion}
This section first shows the result of the TalentPredictor described in the previous section. Then, the behavior, interpretation, and limitation of the model were discussed.

\subsection{Clustering Results}
To evaluate the performance of clustering, Rand Index, and Mutual Information Score \cite{scikit-learn} were used due to their popularity in clustering metrics. Given input dataset $I$ of length $n$, let there be ground truth label $T$ and prediction label $P$ of input data. Let $a$ be the number of pairs of data that have the same label in $T$ and in $P$, and $b$ be the number of pairs of data that have a different label in $T$ and in $P$. Rand index is given by:

\begin{align}
    \text{Rand Index} = \frac{a + b}{\binom{n}{2}}
\end{align}

Then, for the mutual information score, first define entropy for the same set of datasets and labels:

\begin{align}
    H(T) = -\sum^{|T|}_{i=1}P(i)\log (P(i))
\end{align}

Where $|T|$ is the number of different labels in $T$ and $P(i) = \frac{|T_i|}{n}$, where $|T_i|$ is the probability of a random data with label $i$ in $T$. Similar to entropy, the mutual information score is given by:

\begin{align}
    \text{Mutual Information} = \sum_{i=1}^{|T|}\sum_{j=1}^{|P|} P(i,j)\log(\frac{P(i,j)}{P(i)P(j)})
\end{align}

Where $P(i,j) = \frac{|T_i \cap P_j|}{n}$ is the probability of a random data with both label $i$ in $T$ and $j$ in $P$.

Both the Rand Index and Mutual Information indicate the clustering is better with a higher score and max out at 1, which is perfect clustering. Intuitively, they are measuring the similarity between the ground truth clusters and the predicted clusters.

\begin{figure*}[ht]
\centering
\captionsetup[subfloat]{font=footnotesize,textfont=footnotesize,labelfont=footnotesize}
\setkeys{Gin}{width=0.24\linewidth}

\subfloat[Clustering algorithm performance on embedding produced by pre-trained BERT\label{fig:1}]{\includegraphics[width=0.45\linewidth]{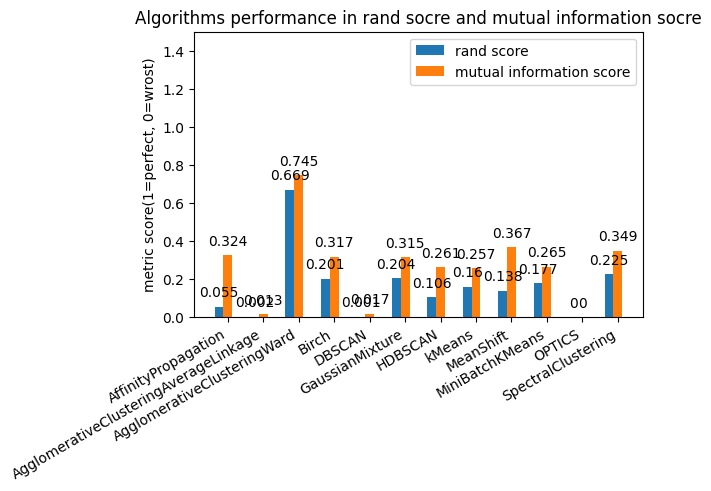} }\hfil
\subfloat[Clustering algorithm performance on embedding produced by fine tuned BERT\label{fig:2}]{\includegraphics[width=0.45\linewidth]{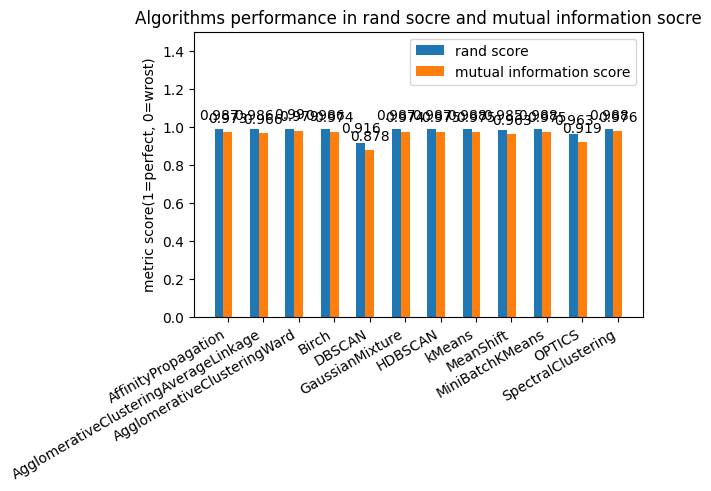}}
\caption{Clustering metrics performance}
\label{fig:plotClusteringmetrics}

\subfloat[Visualization of cluster using pre-trained embedding\label{fig:3}]{\includegraphics[width=0.4\linewidth]{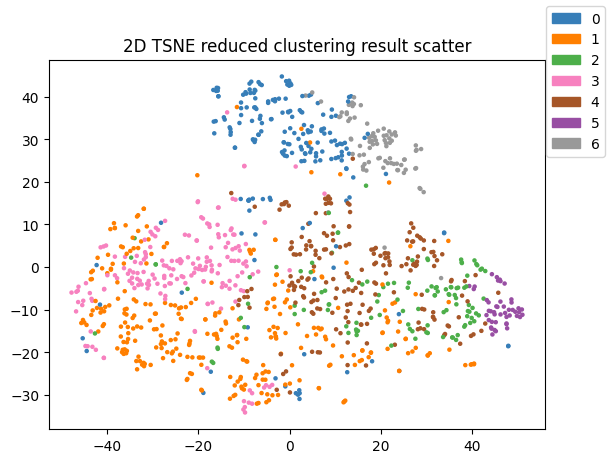} }\hfil
\subfloat[Visualization of cluster using fine tuned embedding\label{fig:4}]{\includegraphics[width=0.4\linewidth]{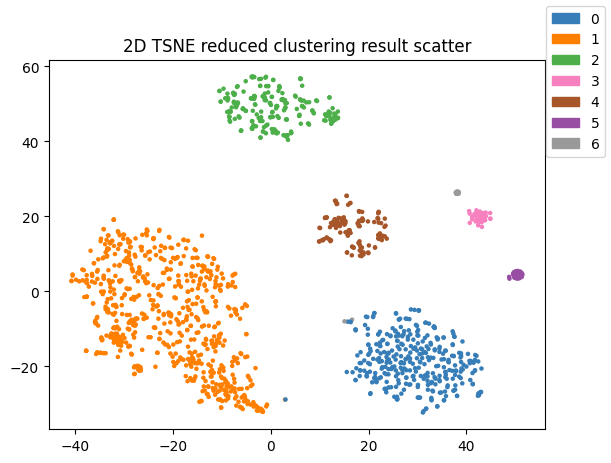}}
\caption{Clustering visualization}
\label{fig:plotClusteringtsne}

\end{figure*}

Figure \ref{fig:1} displays the performance of twelve clustering algorithms on the Rand Index and mutual information score \cite{scikit-learn} for grouping award descriptions into seven clusters using embeddings from pre-trained BERT \cite{cui-etal-2021-pretrain}. Among them, Agglomerative Clustering with Ward linkage outperforms the others significantly.

Figure \ref{fig:2} shows the performance of each of the 12 clustering algorithms on the same task, while the embedding is from fine-tuned BERT. The BERT is fine-tuned using the awards data from 1041 students as training data, and the training task is to classify awards description text into seven talented types. In the fine-tuning, binary cross entropy loss with logit is used with AdamW as the optimizer using default parameter, and Exponential learning rate scheduler is used with gamma of 0.900 \cite{Ansel_PyTorch_2_Faster_2024}. All clustering algorithms perform extremely well on the fine-tuned embedding compared with the pre-trained embedding. All have a Rand Index above 0.900, and most of them reached a Rand Index of 0.970 and a mutual information score of 0.970. While Agglomerative Clustering using Ward linkage still performed the best. Lastly, Table \ref{clusterTab} summarized all clustering performance.

Clustering results using pre-trained embedding and fine-tuned embedding were visualized in Fig. \ref{fig:3} and \ref{fig:4}. Clustering using fine-tuned embedding performed almost perfectly and grouped all embeddings into seven distinguished groups. This shows that fine-tuned BERT can capture the semantics meaning of awards description about which fields it is in nearly perfectly. Thus, using clustering with fine-tuned BERT as an encoder is a valid and accurate auto labler for awards data of Hong Kong secondary school students.

With the accurate auto labeler for ground truth, TalentPredictor is ready to be trained and applied to other secondary schools in Hong Kong. However one limitation is the small size of training and testing data, so it is unknown whether the auto labeler is generalized enough as an accurate auto labeler for all Hong Kong secondary school students. Further research will be carried out to test the generalizability of TalentPredictor in more local schools.

\begin{figure*}[ht]
\centering
\captionsetup[subfloat]{font=footnotesize,textfont=footnotesize,labelfont=footnotesize}
\setkeys{Gin}{width=0.24\linewidth}

\subfloat[Learning curve on binary cross entropy loss with logit and ROCAUC using One (proposed) Encoder\label{fig:5}]{\includegraphics[width=0.4\linewidth]{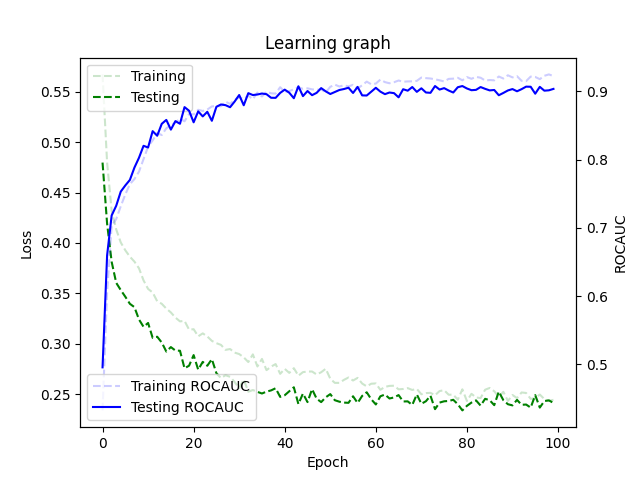} }\hfil
\subfloat[Learning curve on binary cross entropy loss with logit and ROCAUC using Raw Encoder\label{fig:5_5}]{\includegraphics[width=0.4\linewidth]{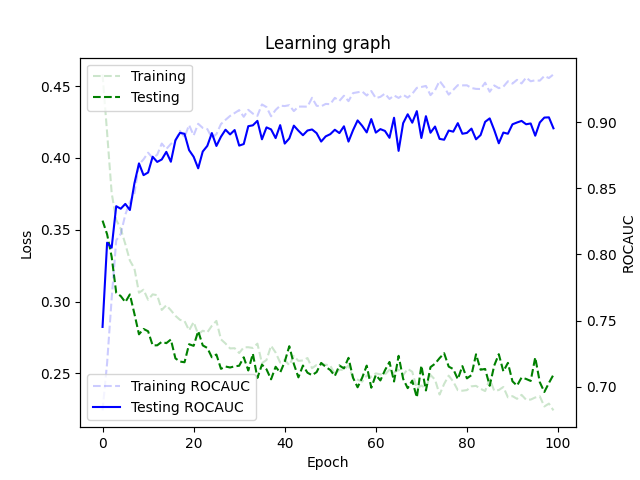}}
\caption{Talented student prediction learning curve}
\label{fig:learning_curve}
\end{figure*}

\begin{table}
\begin{center}
\caption{Clustering Performance on Award Classification}
\label{clusterTab}
\begin{tabular}{| l | c | c | c | c |}
\hline
\multicolumn{1}{|c|}{\multirow{3}{4em}{Clustering Algorithm}} & Pre-train & Pre-train & Finetune & Finetune \\
& Rand & Mutual & Rand & Mutual \\
& Index & Info & Index & Info\\
\hline
OPTICS & 0.000 & 0.000 & 0.963 & 0.919 \\
\hline
DBSCAN & 0.001 & 0.017 & 0.916 & 0.878 \\
\hline
Agglomerative & \multirow{4}{*}{0.002} & \multirow{4}{*}{0.013} & \multirow{4}{*}{0.986} & \multirow{4}{*}{0.966} \\
Clustering &&&&\\
Average &&&&\\
Linkage &&&&\\
\hline
Affinity & \multirow{2}{*}{0.055} & \multirow{2}{*}{0.324} & \multirow{2}{*}{0.987} & \multirow{2}{*}{0.973} \\
Propagation &&&&\\
\hline
HDBSCAN & 0.106 & 0.261 & 0.987 & 0.975 \\
\hline 
MeanShift & 0.138 & 0.367 & 0.985 & 0.963 \\
\hline
K-means & 0.16 & 0.257 & 0.988 & 0.975 \\
\hline
MiniBatch & \multirow{2}{*}{0.177} & \multirow{2}{*}{0.265} & \multirow{2}{*}{0.988} & \multirow{2}{*}{0.975} \\
KMeans &&&&\\
\hline
Birch & 0.201 & 0.317 & 0.986 & 0.974 \\
\hline
Gaussian & \multirow{2}{*}{0.204} & \multirow{2}{*}{0.315} & \multirow{2}{*}{0.987} & \multirow{2}{*}{0.974} \\
Mixture &&&&\\
\hline
Spectral & \multirow{2}{*}{0.225} & \multirow{2}{*}{0.349} & \multirow{2}{*}{0.988} & \multirow{2}{*}{0.976} \\
Clustering &&&&\\
\hline
\textbf{Agglomerative} & \multirow{3}{*}{\textbf{0.669}} & \multirow{3}{*}{\textbf{0.745}} & \multirow{3}{*}{\textbf{0.990}} & \multirow{3}{*}{\textbf{0.979}} \\
\textbf{Clustering} &&&&\\
\textbf{Ward} (Our method) &&&&\\
\hline
\end{tabular}
\end{center}
\end{table}

\begin{table*}
\begin{center}
\caption{Proposed TalentPredictor's performance in prediction}
\label{tab2}
\resizebox{\textwidth}{!}{
\begin{tabular}{| l | c | c | c | c | c | c | c | c |}
\hline 
&\multicolumn{4}{c|}{\textbf{One Encoder} (Proposed method)} & \multicolumn{4}{c|}{Raw Encoder}\\
\hline
\multicolumn{1}{|c|}{Talented} & Train & Test & Train & Test & Train & Test & Train & Test \\
\multicolumn{1}{|c|}{type} & accuracy & accuracy & ROCAUC & ROCAUC & accuracy & accuracy & ROCAUC & ROCAUC \\
\hline
Academic & 0.910 & 0.923 & 0.872 & 0.844 & 0.916 & 0.918 & 0.894 & 0.778 \\
\hline
Sport & 0.842 & 0.841 & 0.896 & 0.844 & 0.861 & 0.851 & 0.924 & 0.869 \\
\hline
Art & 0.904 & 0.928 & 0.886 & 0.897 & 0.912 & 0.937 & 0.913 & 0.878 \\
\hline
Leadership & 0.947 & 0.947 & 0.918 & 0.923 & 0.962 & 0.971 & 0.963 & 0.939 \\

\hline
Service & 0.886 & 0.885 & 0.932 & 0.895 & 0.904 & 0.923 & 0.953 & 0.946 \\
\hline
Technology & 0.964 & 0.976 & 0.900 & 0.918 & 0.972 & 0.976 & 0.946 & 0.773 \\
\hline
Other & 0.821 & 0.793 & 0.884 & 0.853 & 0.826 & 0.784 & 0.887 & 0.851 \\
\hline
Average & 0.896 & 0.899 & 0.898 & \textbf{0.882} & 0.908 & 0.908 & 0.926 & 0.862 \\
\hline
Micro & - & - & 0.925 & 0.908 & - & - & 0.936 & 0.908 \\
\hline
\end{tabular}}
\end{center}
\end{table*}

\subsection{Prediction Results}
To evaluate the prediction, accuracy and the Area Under the Receiver Operating Characteristic Curve (ROCAUC) \cite{scikit-learn} were used. Given input dataset $X$ of lenght $n$, let $y_i$ and $\hat{y}_i$ be the the ground truth and prediction of $i^{\text{th}}$ data $X_i$. With identity function:

\begin{align}
    \mathbb{1}(y_i, \hat{y}_i) = \begin{cases}
        1 & \text{if } y_i = \hat{y}_i,\\
        0 & \text{if } y_i \ne \hat{y}_i,
    \end{cases}
\end{align}

The accuracy is defined as:

\begin{align}
    \text{Accuracy} (y_i, \hat{y}_i) = \frac{1}{n} \sum^n_{i=1} \mathbb{1}(y_i, \hat{y}_i)
\end{align}

As for ROCAUC, firstly consider binary classification for simplicity. Let the number of true positives, true negatives, false positives, and false negatives be TP, TN, FP, and FN respectively. Define true positive rate (TPR) and false positive rate (FPR) as

\begin{align}
    \text{TPR} = \frac{\text{TP}}{\text{TP} + \text{FN}}\\ \text{FPR} = \frac{\text{FP}}{\text{FP} + \text{TN}}
\end{align}

The Receiver Operating Characteristic Curve (ROC) is formed by plotting TPR against FPR, the different TPR and FPR come from adjusting the normalized classification threshold from 0 to 1. ROCAUC is simply the Area under the Curve (AUC) of ROC. Intuitively, ROCAUC represents how well two classes are separable. 
If perfect separation is possible, ROCAUC is 1, the harder it is to separate, the lower the ROCAUC. A value of 0.5 denotes an absence of discriminatory power, equivalent to random assignment, whereas a value of 0 indicates perfect negative discrimination, whereby inverting the class labels would yield a perfect classifier (ROCAUC = 1).

For multi-class scenarios, one-vs-rest approach is used. In it, the ROCAUC of each class is computed against the rest, i.e. each class is positive while the rest is negative, and the final ROCAUC is the micro average of all single-class ROCAUC.

Now, table \ref{tab2} shows the metrics performance of the prediction model for both encoders. The talented student prediction has the best test micro ROCAUC of 0.908 for both encoders, indicating both are excellent classifiers. Also, the average test ROCAUC of One Encoder and Raw Encoder are 0.882 and 0.860 respectively, indicating a very good classifier. However, One Encoder gives a more balanced prediction, shown by the better average test ROCAUC and smaller range for test ROCAUC across all talent types. From \ref{fig:5} and \ref{fig:5_5}, we can also see One Encoder is more stable and less overfitted than the Raw Encoder. Overall, One Encoder is better as it has a higher average test ROCAUC, more balanced, more stable, and less overfitted. Lastly, a possible reason for the One Encoder doing better than the Raw Encoder is the regularization effect of the smaller embedding size of the One Encoder. Basically, the transformer and LSTM are too powerful and are overfitted to training data, and the smaller One Encoder restricts the output of the transformer and LSTM to a reasonable range thus preventing overfitting.

Table \ref{tab2} shows the metrics performance of the prediction model for both encoders. The talented student prediction has the best test micro ROCAUC of 0.908 for both encoders, indicating both are excellent classifiers. Also, the average test ROCAUC of One Encoder and Raw Encoder are 0.882 and 0.860 respectively, indicating a very good classifier. However, One Encoder gives a more balanced prediction, shown by the better average test ROCAUC and smaller range for test ROCAUC across all talent types. From \ref{fig:5} and \ref{fig:5_5}, we can also see One Encoder is more stable and less overfitted than the Raw Encoder. Overall, One Encoder is better as it has a higher average test ROCAUC, more balanced, more stable, and less overfitted. Lastly, a possible reason for the One Encoder doing better than the Raw Encoder is the regularization effect of the smaller embedding size of the One Encoder. Basically, the transformer and LSTM are too powerful and are overfitted to training data, and the smaller One Encoder restricts the output of the transformer and LSTM to a reasonable range thus preventing overfitting.

For metrics, ROCAUC is chosen over more common metrics such as accuracy, precision, and recall. This is because the talented student dataset is heavily unbalanced in both the dataset collected and reality. Any model can easily gain 0.900 or higher in those metrics by setting an extreme threshold of 1 or 0 and blindly giving negative or positive to all input. Hence metrics like ROCAUC measure the separability of the model by considering all possible thresholds as more reliable \cite{harm_of_class_imbalance}. Nonetheless, Table \ref{tab2} showing the metrics performance still used accuracy for easier understanding and comparability. For using ROUAUC, there may be a concern that not all teachers can understand the unpopular metrics and fail to utilize the model in practice. However, ROCAUC is easy to understand. High ROCAUC indicates that the model is more likely to give a positive case higher confidence (on a scale from 0 to 1) than a negative case. Besides, by using ROCAUC, teachers can adaptively label the most talented \(n\) student whom the model gives the top \(n\) confidence as a talented student. The \(n\) students refer to the \(n\) students most likely to be talented by the model. This is much better than using accuracy and fixed threshold, where the teacher can't adaptively select students and has to rely on binary predictions. 

\subsection{The interpretation of TalentPredictor}
Besides model performance on metrics, another aspect of big data in education cares about is the interpretation of the model. However, deep learning is known to be a black box, making it not suitable for interpretation. Despite this, the deep learning model proposed in this research is interpretable. The model uses One Encoder as an automatic features extractor. Thus, the impact of each feature can be interpreted based on the One Encoder embedding representing each feature. Notice that each One Encoder embedding represents one and only one feature, as in the forward passes of TalentPredictor each input feature has no direct or indirect connection to any other features until the final concatenation. So, SHAP (Shapley Additive exPlanations) \cite{NIPS2017_7062} can be used to represent the importance of each feature on the size 1 embedding of One Encoder.

The SHAP values are calculated by using logistic regression as the final classifier with the embeddings before the final concatenation as input and other settings are the same as TalentPredictor. Otherwise, if the original model was used, SHAP would go back into the sequential list for LSTM and the original text of BERT embedding, making the interpretation on each element constructed features instead of on each feature. Meanwhile, logistic regression consumed the embedding just before concatenation. So SHAP will track down the embedding of size one representing each feature instead of every component in the feature, which makes the interpretation clearer and faster. Finally, the prediction performance of using logistic regression as the final classifier is 0.880 and 0.878 for test accuracy and ROCAUC respectively. We can see performance is similar to our best method, hence logistic regression can serve as a good interpretation median of the final embedding. Some may question the performance of the model on other popular classifiers, such as SVM, DT, NB, etc., as they are missing. However, the deep learning model optimized the representation and final classifier together throughout training. So other classifiers not optimized for corresponding representation are not tested for performance. Logistic regression is used here for the sole purpose of interpretation.

\subsection{Ethical Issues}
Here are some ethical concerns about the TalentPredictor model. Since the model focuses only on talented students, there may be concerns about unbalanced resource distribution among students. Teachers might prioritize talented students and overlook at-risk students who require more assistance. To address this concern, an at-risk student detection model has been developed alongside the TalentPredictor model in this research \cite{yang_2024_dmp_ai, cheng2025predicting}. This allows for the identification of talented students while also recognizing at-risk students, assisting teachers in identifying any student who is above or below average, and providing appropriate support.

\section{Conclusion}
This research is the first to explore the possibilities of using a semi-supervised multimodal model in multi-type talented students prediction for offline secondary school education. And is the first to use awards as a feature. The high test micro ROCAUC of 0.908 for talented student prediction shows that the proposed model can distinguish talented students from other students with excellence. 

A limitation of this research is that labeling awards as talented or not still relies on manual annotation. Although the clustering method achieved a Rand index of 0.990, its generalizability to other schools is uncertain. Future work should validate the proposed auto-labeler in more schools and develop better-unsupervised machine learning models for automatic classification. This would eliminate the need for human labeling and create a fully unsupervised talent prediction system. The model has already been deployed in local schools, allowing teachers to use this new talent identification method. With ongoing research and updates, this multimodal analysis approach is set to support secondary school talent identification effectively.

In summary, this research has made significant advancements in leveraging data features, achieving high model performance, and enabling practical application in schools. The innovative use of awards as a predictive signal and the potential for fully unsupervised multi-type talent identification are particularly noteworthy contributions to the field.

\bibliographystyle{IEEEtran}

\end{document}